\documentclass[letterpaper]{article}

\usepackage{ijcai13}
\usepackage{times}
\usepackage[utf8]{inputenc}
\usepackage{local}

\begin{document}

   \title{Case Adaptation with Qualitative Algebras
   }
   \author{
   	Valmi Dufour-Lussier \\
   		LORIA \\
   		Université de Lorraine \\
   		CNRS UMR 7503,
   		Inria \\
   		F-54506 Vand\oe{}uvre \\
   		\texttt{\scriptsize valmi.dufour@loria.fr}
   	\And
   	Florence Le Ber \\
   		ICUBE \\
   		Université de Strasbourg \\
   		ENGEES, CNRS \\
   		F-67000 Strasbourg \\
   		\texttt{\scriptsize florence.leber@engees.unistra.fr}
   	\And
   	Jean Lieber \\
   		LORIA \\
   		Université de Lorraine \\
   		CNRS UMR 7503,
   		Inria \\
   		F-54506 Vand\oe{}uvre \\
   		\texttt{\scriptsize jean.lieber@loria.fr}
   	\And
   	Laura Martin \\
   		INRA \\
   		UR055 \\ ASTER--Mirecourt \\
   		F-88500 Mirecourt \\
   		\!\!\!\!\!\!\texttt{\scriptsize laura.martin@mirecourt.inra.fr}
	}
      
\maketitle
   
   \begin{abstract}
This paper proposes an approach for the adaptation of spatial or temporal cases in a case-based reasoning system.
Qualitative algebras are used as spatial and temporal knowledge representation languages.
The intuition behind this adaptation approach is to apply a substitution and then repair potential inconsistencies,
 thanks to belief revision on qualitative algebras.
A temporal example from the cooking domain is given.\footnote{
     			The paper on which this extended abstract is based was the recipient of the best paper award of the
  	   		2012 International Conference on Case-Based Reasoning~\cite{dufour12iccbr}.}
   \end{abstract}
   
   
   \section{Introduction}
      \label{sec:intro}
Case-based reasoning~(CBR)~\cite{rs89cbr} is a framework in which a new problem (the target case) is solved
 by first retrieving an older, similar problem to which the solution is known (the source case),
 and then adapting this solution to fit the new problem.
While the retrieval stage has been thoroughly studied by the CBR community,
 the adaptation stage has received less attention until recently.
One proposal to address the adaptation problem is to apply a belief revision operator,
 revising source knowledge by target knowledge~\cite{lieber07c}.
In this paper, we apply Lieber's proposal to
 case knowledge represented using a qualitative algebra,
 such as Allen's calculus~\cite{allen83cacm} or RCC8~\cite{rcc92kr}.
 
Qualitative spatial and temporal reasoning (QSTR) as a research domain has been active since the beginning of the 1980s.
The paradigm has been exploited to help solve planning and constraint satisfaction problems,
 but rarely within CBR.
Nevertheless, many domains in which QSTR is used could be addressed with \rapc because the knowledge involved is usually contextual and incompletely formalised.
This is the case in the domain of \AdT, in which knowledge can be acquired from schematic descriptions of the spatial organisation of farmlands.
Another 
 example is the cooking domain,
 in which some knowledge is of a temporal nature.


In section~\ref{sec:motiv}, our approach is illustrated informally using a cooking example.
Section~\ref{sec:bckgr} then introduces the formal notions
 required for the approach, namely in terms of \rapc, revision-based adaptation, and QSTR.
The approach is then defined in details in section~\ref{sec:formalisation}, and an algorithm is described in section~\ref{sec:algo}.
Section~\ref{sec:eval} illustrates those formal notions and the results of the algorithm using the example introduced in section~\ref{sec:motiv}.
Related work is discussed in section~\ref{sec:rw}.

   \section{A cooking example}
      \label{sec:motiv}
\label{sec:motiv-time}

To illustrate temporal case adaptation, we use \taaable~\cite{COJAN-2011-646717}, a CBR application for cooking.
\taaable answers user queries, for instance: ``I want a recipe for a carrot risotto.''
If no matching recipe is found in the cookbook (the case base),
 a recipe of the same type with similar ingredients will be retrieved,
 for instance a mushroom risotto.
\taaable will then suggest the user replaces mushrooms with carrots.
On the other hand, in its current form, it will not be able to help the user in adapting the recipe.

Suppose the mushrooms were added to the rice 2 minutes before the end,
 but the cooking domain knowledge indicates that carrots must be cooked for 25 minutes in order to be done,
 whereas the rice must be cooked for 18 minutes.
A proper adaptation would require not only the lengthening of the cooking time of the vegetables,
 but also a reordering of the actions in the recipe.
Therefore we expect the approach we will now introduce,
 given a retrieved recipe and a requested substitution,
 will be able to reorder the actions of the recipe
 in order to present \taaable users with a usable procedure.

   \section{Background}
      \label{sec:bckgr}

\subsection{Case-based reasoning and case adaptation}
\label{subsec:adaptation}

In this paper, $\Source$, $\Cible$ and $\ConnDom$
 respectively denote the case to be adapted, 
 the target case and the domain knowledge.
$\Source$ and $\Cible$ are required to be consistent with
 $\ConnDom$.
Given $\Source$ and $\Cible$, the adaptation aims at
 building a new case, $\CasAdapte$.
This case is built by adding some information to the target
 case
 (intuitively, $\Cible$ specifies only the ``problem part''
  of the query),
 and it has to be consistent with $\ConnDom$.

It is assumed that a matching step precedes the
 adaptation process, providing links
 between $\Source$ and $\Cible$. It is represented by a substitution
 $\subst$,
mapping descriptors of $\Source$ to descriptors
 of $\Cible$.
As an example, in the system \taaable, matching is performed
 during retrieval.
This process, applied to the cooking example of the previous section, 
 would have returned $\subst=\SUBST{\champignon}{\carotte}$.
In the following, this preprocessing step of adaptation is considered
 to be given and, thus, $\subst$ is an input of the adaptation process
 described in section~\ref{sec:formalisation}.

\subsection{Belief revision and revision-based adaptation}
\label{subsec:revision}

%

In a given representation formalism, a revision operator $\rev$
 maps two knowledge bases $\psi$ and $\mu$ to
 a knowledge base $\psi\rev\mu$, the revision of $\psi$ by $\mu$.
Intuitively, $\psi\rev\mu$ is obtained by making a minimal
 change of $\psi$ into $\psi'$, so that the conjunction of
 $\psi'$ and $\mu$, $\psi'\land\mu$, is consistent.
Then, $\psi\rev\mu$ is this conjunction.

The notion of minimal change can be modelled in various ways,
 so there are various revision operators.
However, postulates have been proposed for such an operator,
 such as the AGM postulates~\cite{agm-85}.
These postulates have been applied to propositional logic~\cite{katsuno91}
 and well studied in this formalism.
Given a distance $\dist$ on the set $\Univers$
 of the interpretations, an operator $\revDist$
 can be uniquely defined (up to logical equivalence) as:
 the set of models of $\psi\revDist\mu$ is the
 set of models of $\mu$ that have a minimal distance to the set
 of models of $\psi$.

%
Given a revision operator $\rev$, $\rev$-adaptation consists simply in
 using this revision operator to perform adaptation, taking into account
 the domain knowledge:
 \begin{equation}
   \CasAdapte = (\ConnDom\land\Source) \rev (\ConnDom\land\Cible)
   \label{eq:adaptation/revision}
 \end{equation}
The intuition behind revision-based adaptation is to reduce
 adaptation to an inconsistency repair.

\subsection{Qualitative representation of temporal knowledge}
\label{subsec:QCN}

\subsubsection{Definitions}
%
A qualitative algebra is a relation algebra that defines a
 set $\ensRelations$ of binary relations applicable between two variables,
 usually representing points, intervals or regions.
Allen interval algebra~\cite{allen83cacm}, for instance,
 introduces $13$ basic relations between intervals, corresponding to
 the $13$ possible arrangements of their lower and
 upper bounds.
$7$ relations are illustrated in figure~\ref{fig:allen}.
The $6$ others are the inverse of the first $6$
 (\relQCNsa{eq} is symmetric).

\indu~\cite{pks99atai} extends the set of Allen relations by combining them with relations over the interval durations. For 7 Allen relations, there is only one possible duration relation (e.g. $i \alrel{d} j$ implies that the duration of $i$ is shorter than the duration of $j$). For the other $6$, all three duration relations $<$, $=$ and $>$ are possible. This yields a total of $25$ basic relations.
They are written as $r^s$, where $r$ is an Allen relation and $s$ is a duration relation. 

\setcounter{figure}{0}

\begin{figure}
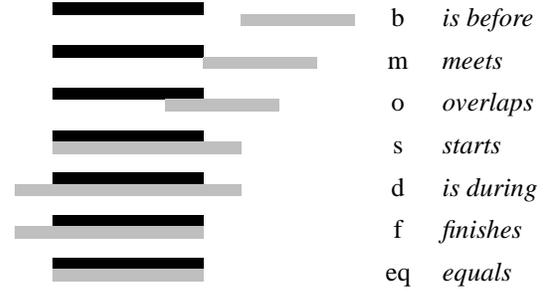

	\centering
{
  \def\intx{$\rule[1.5mm]{20mm}{1.5mm}$}
  \def\inty#1#2{{\textcolor{gray!50}{\hspace{#1mm}$\rule[0mm]{#2mm}{1.5mm}$}}}
  \def\ligne#1#2#3#4{\makebox[0mm][l]{\intx}\inty{#1}{#2} & $\alrelnb{#3}$ & \emph{#4}}
  \def\rien#1{#1} 
  %
    \begin{tabular}{l c l}
      \ligne{25}{15}{b}{is \rien{b}efore}
      \\[1.5mm]
      \ligne{20}{15}{m}{\rien{m}eets}
      \\[1.5mm]
      \ligne{15}{15}{o}{\rien{o}verlaps}
      \\[1.5mm]
      \ligne{0}{25}{s}{\rien{s}tarts}
      \\[1.5mm]
      \ligne{-5}{30}{d}{is \rien{d}uring}
      \\[1.5mm]
      \ligne{-5}{25}{f}{\rien{f}inishes}
      \\[1.5mm]
      \ligne{0}{20}{eq}{\rien{eq}uals}
    \end{tabular}
  %
}

	\caption{\label{fig:algebras}\label{fig:allen}Allen interval algebra basic relations.}
\end{figure}


Qualitative knowledge can be represented as qualitative constraint networks (QCNs).
A QCN is a pair $(V,C)$, where $V$ is a set of variables, and $C$ is a set of
 constraints of the form \contrainteQCNsa{V_i}{V_j}{C_{ij}} with $V_i, V_j \in V$,
 and $C_{ij}$ is a set of the basic relations defined by the algebra
 ($C_{ij}$ is a relation that is a disjunction of the basic relations,
  i.e. $i\mathrel{\{r_1, r_2\}}j$ means that $i$ is
  related to $j$ with either $r_1$ or $r_2$).
In \indu, shortcut notations $r^?$ and $?^s$ respectively
 represent the Cartesian product of $r$ and all possible
 duration relations and the product of $s$ and all possible
 Allen relations
 (e.g.,
  $\relQCNindu{m}{$?$}
   =
   \{\relQCNsaindu{m}{$<$},\relQCNsaindu{m}{$=$},\relQCNsaindu{m}{$>$}\}$;
  $\relQCNindu{d}{$?$}
   =
   \{\relQCNsaindu{d}{$<$}\}$;
  $\relQCNindu{?}{$=$}
   =
   \{\relQCNsaindu{b}{$=$},\relQCNsaindu{m}{$=$},\relQCNsaindu{o}{$=$},\relQCNsaindu{eq}{$=$},\relQCNsaindu{oi}{$=$},\relQCNsaindu{mi}{$=$},\relQCNsaindu{bi}{$=$}\}$).

A \emph{scenario} is a QCN $\scenario=(V_\scenario, C_\scenario)$
 such that for each $V_i, V_j\in{}V_\scenario$, there exists one
 constraint $\contrainteQCN{V_i}{V_j}{$r$}\in{}C_\scenario$.
$\scenario$
 satisfies the QCN $\QCN = (V_\QCN, C_\QCN)$
 if $\scenario$ and $\QCN$ have the same set of variables and each 
 constraint relation in $\scenario$ is a subset of the corresponding constraint
 relation in $\QCN$.
A scenario is consistent if a valuation can be provided for the variables
 such that all constraints are observed, and a QCN is consistent
 if it has a consistent scenario.
Two QCNs are said to be equivalent if every scenario of the former is a scenario
 of the latter and vice-versa.

\subsubsection{Revision of QCNs}
%
A QCN is a knowledge base and thus, the issue of revising a QCN
 $\psi$ by a QCN $\mu$ can be addressed.
Building on the work of~\cite{schwind08flairs}, we defined
 a revision operator for QCNs,
 following the idea of an operator $\revDist$
 (cf. section~\ref{subsec:revision}),
 where an interpretation is a scenario, a model of a QCN is
 a scenario that satisfies it, and a distance $\dist$ between
 scenarios/interpretations is defined as follows.

First, a distance $d$ between basic relations of the considered
 algebra is defined.
Formally, a \emph{neighbourhood graph} whose vertices are the 
 relations of the algebra is given, and $d(r, s)$
 is the distance between $r$ and $s$ in the
 graph.
It represents closeness between relations.
For instance, $\relQCNsa{b}$ and $\relQCNsa{m}$ are close
 \mbox{($d(\relQCNsa{b}, \relQCNsa{m})=1$)}
 since they express similar conditions on the boundaries of the
 intervals
 (for the lower bounds: $=$ for both;
  for the upper bounds: $<$ for $\relQCNsa{b}$ and $=$ for
  $\relQCNsa{m}$).
$d$ makes it possible to define $\dist$, a distance between two scenarios
 $\scenarioS=(V, C_{\scenarioS})$ and $\scenarioT=(V, C_{\scenarioT})$
 based on the same set of variables $V$\!, as:
 \begin{equation}
 \label{eq:scenario-distance}
   \dist(\scenarioS, \scenarioT)
   =
   \sum_{V_i, V_j\in{}V, i\neq{}j}
   d(r_\scenarioS(V_i, V_j), r_\scenarioT(V_i, V_j))
 \end{equation}
 where $r_\scenarioS(V_i, V_j)$
 is the relation $r$ such that $V_i\mathrel{\{r\}}V_j\in{}C_\scenarioS$.

Given two QCNs $\psi$ and $\mu$, the revision of $\psi$ by
 $\mu$ returns the set $R$ of scenarios satisfying $\mu$ that are the
 closest ones to the set of scenarios satisfying $\psi$.\footnote{%
   This slightly differs from the definition of revision given in
    section~\ref{subsec:revision} where $\psi\rev\mu$ is a knowledge
    base, not a set of models.
}
\section{Formalisation}
\label{sec:formalisation}

\subsection{Representation of the adaptation problem}

\subsubsection{Parametrised QCNs}
%
It is assumed that the variables of the considered QCNs can
 be parametrised by elements of a given set $\Parametres$.
A parameter $\param\in\Parametres$ is
 either a \emph{concrete parameter}, $\param\in\ParametresConcrets$,
 or an \emph{abstract parameter}, $\param\in\ParametresAbstraits$:
 $\Parametres=\ParametresConcrets\cup\ParametresAbstraits$,
 $\ParametresConcrets\cap\ParametresAbstraits=\emptyset$.
A concrete parameter denotes a concept of the application domain,
 e.g. $\champignon\in\ParametresConcrets$ for the cooking example.
In this example, the formal interval $\cuire(\champignon)$
 represents the temporal interval of the mushroom cooking.
The domain knowledge $\ConnDom=(V_\ConnDom, C_\ConnDom)$
 is a set of constraints, for example:
 \begin{equation*}
   C_\ConnDom=
   \left\{
   \text{%
     \begin{tabular}{l}
       $\cuire(\riz) \releqduree \XVIIImin$\\
       $\cuire(x) \relm \cuit(x)$\\
       $\XVIIImin \relinfduree \XXVmin$
     \end{tabular}
   }
   \right\}
   \label{eq:ex-CD}
 \end{equation*}
 where $\riz\in\ParametresConcrets$ and $x\in\ParametresAbstraits$,
 represents the facts that rice requires 18~minutes of cooking,
 that $x$ is cooked as soon as the action of cooking $x$ is finished,
 and that 18~minutes are shorter than 25~minutes.
An abstract parameter must be understood with a universal
 quantification over the concrete parameters; e.g.
 $\cuire(x) \relm \cuit(x)$ entails
 $\cuire(\champignon) \relm \cuit(\champignon)$.

Let $\QCNN1$ and $\QCNN2$ be two QCNs.
$\QCNN1\land\QCNN2$ is the QCN \mbox{$\QCN=(V, C)$}
 such that $V=V_1\cup{}V_2$ and $C$ contains the constraints of
 $C_1$, the constraints of $C_2$, and the constraints that are
 deduced by instantiation of the abstract parameters by concrete
 parameters appearing in $\QCNN1$ and $\QCNN2$
For example, if $\QCNN1=C_\ConnDom$ defined by
 equation~(\ref{eq:ex-CD}) and
 $\QCNN2=(\{\cuire(\tomate), \cuit(\tomate)\}, \emptyset)$,
 then $\QCNN1\land\QCNN2=(V, C)$ with
 $C=C_\ConnDom\cup\{\cuire(\tomate) \relm \cuit(\tomate)\}$.


\subsubsection{Substitutions}
%
The \emph{atomic substitution} $\subst=\SUBST{\paramP}{\paramQ}$,
 where $\paramP, \paramQ\in\Parametres$,
 is the function from $\Parametres$ to $\Parametres$
 defined by
 $\subst(\paramA)=\begin{cases}
                  q & \text{if }\paramA=\paramP \\
                  \paramA &\text{otherwise}
                  \end{cases}$.
A \emph{substitution} is a composition
 $\subst_{1}\comp\ldots\comp\subst_n$
 of atomic substitutions $\subst_i$.

Let $\subst=\SUBST{\paramP}{\paramQ}$ be an atomic substitution.
$\subst$ is \emph{concrete} if $\paramP, \paramQ\in\ParametresConcrets$.
$\subst$ is an \emph{atomic abstraction} if
 $\paramP\in\ParametresConcrets$ and $\paramQ\in\ParametresAbstraits$.
$\subst$ is an \emph{atomic refinement} if
 $\paramP\in\ParametresAbstraits$ and $\paramQ\in\ParametresConcrets$.
A \emph{concrete substitution}
 (resp., an abstraction, a refinement)
 is a composition of concrete atomic substitutions
 (resp., of atomic abstractions, of atomic refinements).
Any concrete substitution $\subst$
 can be written $\subst=\abstraction\comp\raffinement$
 where $\abstraction$ is an abstraction and $\raffinement$
 is a refinement, as the following equation 
 illustrates:
 \begin{equation*}
   \SUBST{\champignon}{\carotte}
   =
   \SUBST{\champignon}{x}
   \comp
   \SUBST{x}{\carotte}
 \end{equation*}
 where $\champignon, \carotte\in\ParametresConcrets$
 and $x\in\ParametresAbstraits$.
This can be shown as follows.
First, $\subst$ can be written
 $\SUBST{\paramPN1}{\paramQN1}\comp\ldots\comp\SUBST{\paramPN{n}}{\paramQN{n}}$
 with $\paramPN{i}, \paramQN{i}\in\ParametresConcrets$
 and $\paramPN{i}\neq\paramPN{j}$ if $i\neq{}j$.
Let $x_1$, \ldots, $x_n$ be $n$ abstract parameters,
 let $\abstractionN{i}=\SUBST{\paramPN{i}}{\paramXN{i}}$,
 let $\raffinementN{i}=\SUBST{\paramXN{i}}{\paramQN{i}}$,
 let $\abstraction=\abstractionN1\comp\ldots\comp\abstractionN{n}$,
 and let $\raffinement=\raffinementN1\comp\ldots\comp\raffinementN{n}$.
$\abstraction$ is an abstraction, $\raffinement$ is a refinement
 and $\subst=\abstraction\comp\raffinement$.

Let $\subst$ be a substitution.
$\subst$ is extended on qualitative variables by applying it to
 their parameters.
For example, if $\subst=\SUBST{\champignon}{\carotte}$
 then $\subst(\cuire(\champignon))=\cuire(\carotte)$.
Then, $\subst$ is extended to a constraint
 $c=(V_i\mathrel{C_{ij}}V_j)$ by
 $\subst(c)=(\subst(V_i)\mathrel{C_{ij}}\subst(V_j))$.
Finally, $\subst$ is extended on a QCN by applying it to its
 variables and constraints:
 $\subst((V, C))=(\subst(V), \subst(C))$
 where $\subst(V)=\{\subst(V_i)~|~V_i\in{}V\}$
 and $\subst(C)=\{\subst(c)~|~c\in{}C\}$.

\subsubsection{Adaptation problem}
%
An adaptation problem is given by a tuple\linebreak
 $(\Source, \Cible, \ConnDom, \subst)$.
$\Source$ and $\Cible$ are the representations of the source
 and target cases by QCNs with concrete variables
 (i.e. not parametrised by any abstract parameter).
 $\ConnDom$ is a QCN representing the domain knowledge.
$\subst = \SUBST{\paramPN1}{\paramQN1}\comp\ldots\comp\SUBST{\paramPN{n}}{\paramQN{n}}$
 is a concrete substitution such that each $\paramPN{i}$ (resp., $\paramQN{i}$)
 parametrises a variable of $\Source$ (resp., $\Cible$).
$\ConnDom\land\Source$ and $\ConnDom\land\Cible$ are assumed to be consistent
 (cf. section~\ref{subsec:adaptation}).
The goal of adaptation is to build a consistent QCN $\CasAdapte$
 that entails $\ConnDom\land\Cible$, whose qualitative variables are
 obtained by applying $\subst$ on the qualitative variables of $\Source$,
 and that is obtained thanks to minimal modification of
 $\ConnDom\land\Source$.

\subsection{Principles of revision-based adaptation of a QCN}

A first idea to perform the adaptation, given a tuple
 $(\Source, \Cible, \ConnDom, \subst)$, is to apply $\subst$ on
 $\Source$, thus obtaining a QCN $\ConnDom\land\subst(\Source)$
 that may be inconsistent, and then restoring consistency.
Although this gives a good intuition of the revision-based
 adaptation of a QCN, it is not consistent with the
 irrelevance of syntax principle.
(Indeed, any two inconsistent knowledge bases
 are equivalent:
 their sets of models are both empty.)
Thus, at a semantic level,
 repairing
 an inconsistent knowledge base is meaningless.
By contrast, revision aims at modifying a \emph{consistent}
 knowledge base with another \emph{consistent} one,
 the conjunction of which may be inconsistent.

Therefore,
 the revision-based adaptation consists first in decomposing
 $\subst$ in an abstraction $\abstraction$ and a refinement
 $\raffinement$: $\subst=\abstraction\comp\raffinement$.
Then, $\abstraction$ is applied to $\Source$:
 a QCN $\ConnDom\land\abstraction(\Source)$ is built that is
 necessarily consistent since $\ConnDom\land\Source$ is consistent
 and every constraint of $\ConnDom\land\abstraction(\Source)$ corresponds
 to a constraint of $\ConnDom\land\Source$.
In other words, $\ConnDom\land\Source$ is consistent and is more
 or equally constrained as $\ConnDom\land\abstraction(\Source)$,
 so $\ConnDom\land\abstraction(\Source)$ is consistent.

The third step involves revision.
 The idea is to make a revision of $\psi$ by $\mu$ where
 $\psi = \ConnDom \land \abstraction(\Source)$ and
  $\mu=\ConnDom\land\Cible\land\QCNraffinement$
 where $\QCNraffinement$ represents the following statement:
 ``Each qualitative variable $V_i$ of $\abstraction(\Source)$
    is constrained to be equal to its refinement $\raffinement(V_i)$.''
For this purpose, the relation $\egaliteAQ$ for equality
 is used (${\relQCNsaindu{eq}{=}}$ in \indu):
 $V_i \egaliteAQ \raffinement(V_i)$.
 Therefore, $\QCNraffinement = (V_\raffinement, C_\raffinement)$ where
 \begin{align*}
 	V_{\raffinement} =
        \abstraction(V) \cup \subst(V) \quad;
 	\quad
 	C_{\raffinement} =
        \{ V_i \egaliteAQ \raffinement(V_i) ~|~ V_i\in\abstraction(V)\}
 \end{align*}
$\mu$ is consistent since $\ConnDom\land\Cible$ is and
  since each constraint $V_i \egaliteAQ \raffinement(V_i)$ of
  $\QCNraffinement$ either is a tautology
  (when $V_i$ does not contain any abstract parameter refined by $\raffinement$)
  or links a variable $V_i$ that does not appear in
  $\ConnDom\land\Cible$ with $\raffinement(V_i)$.

Then, $\psi\rev\mu$ gives a set of scenarios and $\CasAdapte$
 is chosen among them.


\section{Algorithm and implementation}
\label{sec:algo}
The revision algorithm takes as input $\psi=\ConnDom\land\abstraction(\Source)$,
$\mu=\ConnDom\land\Cible\land \QCNraffinement$,
as well as a relation neighbourhood graph and a transitivity table
 for the algebra used.
The neighbourhood graph enables to define a distance $d$ between relations
 and the transitivity table defines a relation composition function
 ${\compositionDeRelations}:\ensRelations\times\ensRelations\to2^\ensRelations$,
 for example,
 ${\relQCNsa{m}}\compositionDeRelations{\relQCNsa{mi}}
  ={\relQCN{eq, f, fi}}$ in Allen algebra.

First, it is necessary to ensure  that all variables in either QCN are
present in the other QCN as  well.
All pairs of variables that have no relation associated to them  are
 given the relation \ensRelations--the unspecified relation.

The algorithm  must then search within the scenarios of $\mu$ the ones that minimise the distance to $\psi$.
The distance between  the QCNs $\psi$  and $\mu$ is the  smallest distance
 between any scenario of $\psi$ and a scenario $\mu$,
 computed using equation~(\ref{eq:scenario-distance}).
Considering that the minimum of sums is never less than the sum of minimums,
 a lower bound on the distance between two QCNs can be obtained in time $O(|V|^2\cdot|\ensRelations|^2)$
 by computing the pair-wise minimal distance for each constraint and summing those.
This defines an admissible heuristic which is used to instantiate an \Astar search.
The initial state is $\mu$ and a goal state is a scenario of $\mu$.
A successor state is obtained by selecting one constraint and keeping only one relation on this constraint.
The QCN $\psi$ is used in the cost in the heuristic functions.

The amount of scenarios for a QCN is of the order of
 $O\left(|\ensRelations|^{\frac{|V|\cdot(|V|-1)}{2}}\right)$.

   \section{Result}
      \label{sec:eval}
This section revisits the example from section~\ref{sec:motiv}.

Most temporal aspects of recipes can be represented in \indu by reifying cooking actions, ingredient states, and durations as intervals.
For instance, the following could be included in the domain knowledge:
	$\cuire(\carotte)\relQCNindu{m}{?}\cuit(\carotte)$ and
	$\cuire(\carotte)\releqduree\dureeindiquee{25}$, with the provision that, e.g.
	$\dureeindiquee{18}\relinfduree\dureeindiquee{25}$.

To limit the amount of variables shown, we simplify the representation by replacing duration intervals
 with duration relations between the relevant action intervals.
In this representation, $\psi$ contains
\begin{equation*}
	\begin{array}{l}
        C_\ConnDom = \left\{
   	\begin{array}{l}
   		\cuire(\riz) \relinfduree \cuire(\carotte) \\
   		\cuire(\riz) \relm \servir \\
   		\cuire(\carotte) \relm \servir \\
   	\end{array}
   	\right\} \\
   	C_{\abstraction(\Source)} = \left\{
			\cuire(x) \relQCNbaseindu{f}{<} \cuire(\riz)
   	\right\}
	\end{array}
\end{equation*}
In \taaable, there is no firm adaptation constraint from $\Cible$
 ($C_\Cible=\emptyset$)
 therefore $\mu$ contains simply the constraints
\begin{equation*}
        \begin{array}{rl}
   	C_\ConnDom &= \left\{
   	\begin{array}{l}
   		\cuire(\riz) \relinfduree \cuire(\carotte) \\
   		\cuire(\riz) \relm \servir \\
   		\cuire(\carotte) \relm \servir \\
   	\end{array}
   	\right\} \\
		C_\raffinement &= \{ \cuire(x) \releqduree \cuire(\carotte) \}
	\end{array}
\end{equation*}

The revision algorithm returns two scenarios which are predictably distinguished only by the duration relation between $\servir$ and the other actions, since this relation is defined as being unimportant in the domain knowledge. One scenario
 $\scenarioT=(V_\scenarioT, C_\scenarioT)$ is such that $C_\scenarioT$ is
\begin{equation*}
	\left\{
	\begin{array}{l}
		\cuire(x) \relQCNbaseindu{m}{>} \servir \\
		\cuire(\carotte) \relQCNbaseindu{m}{>} \servir, \quad \\
		\cuire(\riz) \relQCNbaseindu{m}{>} \servir \\ 
		\cuire(x) \relQCNbaseindu{eq}{=} \cuire(\carotte) \\
		\cuire(x) \relQCNbaseindu{fi}{>} \cuire(\riz) \\
		\cuire(\carotte) \relQCNbaseindu{fi}{>} \cuire(\riz) \\
	\end{array}
	\right\}
\end{equation*}

In both scenarios, the lengthening of the vegetable cooking is associated with the inversion of the relation between the vegetable and the rice,
i.e. \relQCNsaindu{f}{$<$} becomes \relQCNsaindu{fi}{$>$},
which corresponds to the expected order inversion between the start of both actions.
Therefore, the adaptation is successful.

   \section{Related work}
      \label{sec:rw}
Several research work focused on the representation of time within the  
CBR framework. Most  were interested in the analysis or in the  
prediction of temporal processes
(e.g. breakdown or disease diagnosis starting from regular observations
or successive events).
The temporal aspect is generally taken into account from sequences of events
or sometimes from relative or absolute time stamps~\cite{dojat98,ma03,sanchez05}.
Particularly, the problem of temporal adaptation has been given much attention
 in CBR with a workflow representation~\cite{minor10iccbr}.
Only a few work~\cite{jaczynski98,jaere02} adopted a qualitative representation.
In~\cite{jaere02}, cases are represented by temporal graphs and the  
retrieval step is based on graph matching.
In~\cite{jaczynski98}, cases are indexed by chronicles and temporal  
constraints, which are
represented with a subset of Allen relations.

Some recent work also dealt with a combination of CBR and spatial reasoning,
 for instance in order to improve web services for spatial
 information~\cite{osman06},
 or for spatial event prediction in hostile territories~\cite{li-iccbr09}.

   \section{Conclusion}
      \label{sec:concl}
Qualitative algebras are important to the
 field of knowledge representation and are especially useful
 for qualitative reasoning on space and on time, but their
 use in \rapc has received very little attention
 so far.
This paper focuses on the adaptation of cases represented
 in a qualitative algebra. A cooking example uses the
 temporal algebra $\indu$.
This adaptation uses the principles of revision-based adaptation
 and combines it with a matching between the source and
 target cases.

A prototype for adaptation of cases represented in a qualitative
 algebra has been implemented in Perl\footnote{
 	The Perl library and Java bindings for this and other revision tools are available at \texttt{http://revisor.loria.fr}.
 }
 and applied to the examples
 of this paper, but it is time-consuming and still requires
 improvements in order to be integrated into an operational system
 like \taaable.

	\bibliographystyle{named} 
   \bibliography{./biblio}

\end{document}